# CORL: A Continuous-state Offset-dynamics Reinforcement Learner

**Emma Brunskill**\*    **Bethany R. Leffler**†    **Lihong Li**†    **Michael L. Littman**†    **Nicholas Roy**\*

\* Computer Science and Artificial Intelligence Laboratory
Massachusetts Institute of Technology
Cambridge, MA 02143

† Department of Computer Science
Rutgers University
Piscataway, NJ 08854

## Abstract

Continuous state spaces and stochastic, switching dynamics characterize a number of rich, real-world domains, such as robot navigation across varying terrain. We describe a reinforcement-learning algorithm for learning in these domains and prove for certain environments the algorithm is probably approximately correct with a sample complexity that scales polynomially with the state-space dimension. Unfortunately, no optimal planning techniques exist in general for such problems; instead we use fitted value iteration to solve the learned MDP, and include the error due to approximate planning in our bounds. Finally, we report an experiment using a robotic car driving over varying terrain to demonstrate that these dynamics representations adequately capture real-world dynamics and that our algorithm can be used to efficiently solve such problems.

## 1 INTRODUCTION

Reinforcement learning (RL) has had some impressive successes, such as model helicopter flying (Ng et al., 2004) and expert software backgammon players (Tesauro, 1994). Two key challenges in reinforcement learning are scaling to large worlds, which often involves a form of generalization, and efficiently handling the exploration/exploitation tradeoff. Many real-life problems involve real-valued state variables: discretizing such environments causes an exponential growth in the number of states as the state dimensionality increases, and so solutions that directly reason with continuous-states are of important consideration.

In this paper, we build on recent work on probably efficient reinforcement learning (Kearns & Singh, 2002; Brafman & Tennenholtz, 2002; Strehl et al., 2006) and focus on continuous-state, discrete-action environments. We consider the case when the dynamics can be described as switching noisy offsets where the parameters of the dynamics depend on the state's "type" $t$ and the action taken $a$. More formally,

$$s' = s + \beta_{at} + \varepsilon_{at}, \qquad (1)$$

where $s$ is the current state, $s'$ is the next state, $\varepsilon_{at} \sim \mathcal{N}(0, \Sigma_{at})$ is drawn from a zero-mean Gaussian with covariance $\Sigma_{at}$ and $\beta_{at}$ is the offset.

An example where we expect such dynamics to arise is during autonomous traversal of varying terrain. Here, types represent the ground surface, such as dirt or rocks. The dynamics of the car may be approximated by an offset from the prior state plus some noise, where the offset and noise depend on the surface underneath the car. These models could be useful approximations in a number of other problems, including transportation planning (learning the mean speed and variance of interstate highways and local streets for path planning to a goal location), and packet routing (learning that wireless and ethernet have different bandwidth/usage patterns and routing accordingly).

We present a new RL algorithm for learning in continuous-state, discrete-action Markov decision processes (MDPs) with switching noisy offset dynamics and show that this algorithm is probably approximately correct (PAC) in certain environments with a sample complexity that scales polynomially with the state space dimensionality. We perform planning using fitted value iteration (FVI) and incorporate the error due to approximate planning into our bounds.

Finally, we present experiments on a small robot task that involves navigation over varying terrains. These experiments demonstrate that our dynamics models can adequately capture real-life dynamics and our algorithm can quickly learn good policies in such environments.

## 2 A CONTINUOUS-STATE OFFSET-DYNAMICS REINFORCEMENT LEARNER

This section introduces terminology and then presents our algorithm.

## 2.1 BACKGROUND

The world is characterized by a continuous-state discounted MDP $M = \langle S, A, p(s'|s,a), R, \gamma \rangle$ where $S \subseteq \mathbb{R}^{N_{dim}}$ is the $N_{dim}$-dimensional state space, $A$ is a set of discrete actions, $p(s'|s,a)$ is the unknown transition dynamics that satisfy the parametric form of Equation 1, $\gamma \in [0, 1)$ is the discount factor and $R : S \times A \to [0, 1]$ is the known reward model. In addition to the standard MDP formulation, each state $s$ is associated with a single observable type $t \in T$. The total number of types is $N_T$. The dynamics of the environment are determined by the current state type $t$ and action $a$ taken:

$$p(s'|s,a) = \mathcal{N}(s'|s + \beta_{at}, \Sigma_{at}). \quad (2)$$

In other words, types partition the state space into regions, and each region is associated with particular pair of dynamics parameters.

In this work, we focus on the known reward model, unknown dynamics model situation. The parameters of the dynamics model, $\beta_{at}$ and $\Sigma_{at}$, are assumed to be unknown for all types $t$ and actions $a$ at the start of learning. This model is a departure from prior related work (Abbeel & Ng, 2005; Strehl & Littman, 2008), which focuses on a more general linear dynamics model but assumes a single type and that the variance of the noise $\Sigma_{at}$ is known. We argue there exist interesting problems where the variance of the noise is unknown and estimating this noise may provide the key distinction between the dynamics models of different types.

In reinforcement learning, the agent must learn to select an action $a$ based on its current state $s$. At each time step, it receives an immediate reward $r$ also based on its current state[1]. The agent then moves to a next state $s'$ according to the dynamics model. The goal is to learn a policy $\pi : S \to A$ that allows the agent to choose actions to maximize the total rewards it receives. The value of a particular policy is the expected discounted sum of future rewards that will be received from following this policy, and is denoted $V^\pi(s) = E_\pi[\sum_{j=0}^{\infty} \gamma^j r_j | s_0 = s]$, where $r_j$ is the reward received on the $j$-th time step and $s_0$ is the initial state of the agent. Let $\pi^*$ be the optimal policy, and its associated value function be $V^*(s)$.

## 2.2 ALGORITHM

Our algorithm (*c.f.*, Algorithm 1) is derived from the R-max algorithm of Brafman and Tennenholtz (2002). We first form a set of $\langle t, a \rangle$ tuples, one for each type–action pair. Note that each tuple corresponds to a particular pair of dynamics model parameters, $\langle \beta_{at}, \Sigma_{at} \rangle$. A tuple is considered to be "known" if the agent has been in type $t$ and

---

**Algorithm 1** CORL
1: **Input:** $N_A, N_{dim}, N_T, R, \Sigma_{\max}, \Sigma_{\min}, \gamma, \epsilon$, and $\delta$.
2: Set all type–action tuples $\langle t, a \rangle$ to be unknown and initialize the dynamics models (see text) to create an empirical known-state MDP model $\hat{M}_K$.
3: Select a fixed set of evenly spaced points for fitted value iteration.
4: Start in a state $s_0$.
5: **loop**
6:    Solve MDP $\hat{M}_K$ using fitted value iteration and denote its optimal value function by $Q_t$.
7:    Select action $a = \arg\max_a Q_t(s, a)$.
8:    Transition to the next state $s'$.
9:    Increment the appropriate $n_{at}$ count (where $t$ is the type of state $s$) given the observed transition tuple $\langle s, a, s' \rangle$.
10:   If $n_{at}$ exceeds $N_{at}$ where $N_{at}$ is specified according to the analysis, then mark $\langle a, t \rangle$ as "known" and estimate the dynamics model parameters for this tuple.
11: **end loop**

---

taken action $a$ a number $N_{at}$ times. At each timestep, we construct a new MDP $\hat{M}_K$ as follows. If the number of times a tuple has been experienced, $n_{at}$, is greater than or equal to $N_{at}$, then we estimate the parameters for this dynamics model using maximum-likelihood estimation:

$$\tilde{\beta}_{at} = \frac{\sum_{i=1}^{n_{at}} (s'_i - s_i)}{n_{at}} \quad (3)$$

$$\tilde{\Sigma}_{at} = \frac{\sum_{i=1}^{n_{at}} (s'_i - s_i - \tilde{\beta}_{at})(s'_i - s_i - \tilde{\beta}_{at})^T}{n_{at}} \quad (4)$$

where the sum ranges over all state action pairs experienced for which the type of $s_i$ was $t$ and the action taken was $a$.

Otherwise, we set the dynamics model for all states and action associated with this type–action tuple to be a transition with probability 1 back to the same state. We also modify the reward function for all states associated with an unknown type–action tuple $\langle t_u, a_u \rangle$ so that all state–action values $Q(s_{t_u}, a_u)$ have a reward of $V_{\max}$ (the maximum value possible, $1/(1-\gamma)$). We then seek to solve $\hat{M}_K$. This MDP includes switching dynamics with continuous states, and we are aware of no exact optimal planners for such MDPs[2]. Instead, we will use fitted value iteration to approximately solve the MDP.

In FVI, the value function is represented explicitly at only a fixed set of states that are (for example) uniformly spaced in a grid over the state space. Planning requires performing Bellman backups for each grid point $\mu_f$. Since we are *only* performing backups of the value function at a set of grid points $\mu_f$, we need a function approximator to estimate the value of other points that are not in this fixed set. We can use Gaussian kernel functions to interpolate the value at the grid points to other points. The value of a state $s$ is

$$V(s) = \max_a \sum_{f=1}^{F} w_f \mathcal{N}(s; \mu_f, \Sigma_f) Q(\mu_f, a), \quad (5)$$

where $w_f$ is a scalar and $\mathcal{N}(s; \mu_f, \Sigma_f)$ represents a Gaussian with mean at grid point $\mu_f$ and variance $\Sigma_f$ evaluated at state $s$. The grid-point locations, variances and weights $(\mu_f, \Sigma_f, w_f)$ are defined so

$$\sum_{f=1}^{F} w_f \mathcal{N}(s; \mu_f, \Sigma_f) \approx 1 \quad (6)$$

for all states $s$ of interest. We would like this expression to exactly equal 1 for all states of interest as that guarantees the function approximator is an averager and therefore discounted infinite horizon fitted value iteration is guaranteed to converge (Gordon, 1995). In practice, if Gaussians are placed at uniform intervals over the state space of interest, then this expression can be extremely close to 1. Indeed, as long as the sum in Equation 6 sums to less than or equal to 1 for all states, then the approximator operator is guaranteed to be a non-expansion in the max norm and therefore discounted infinite horizon fitted value iteration is still guaranteed to converge.

---

[1] For simplicity, the reward is assumed to be only a function of state in this paper, but the arguments can be easily extended to where the reward model is also a function of the action chosen.

[2] In contrast, optimal control is possible for linear Gaussian (non-switching) dynamics continuous-state systems with linear quadratic reward functions.

Substituting this representation of the value function in place of $V(s')$ and using the dynamics model in the Bellman backup equation, we can perform the integration over future reward in closed form to get

$$V(\mu.) = R(\mu.) + \gamma \max_a \sum_f w_f \cdot$$
$$\mathcal{N}(\mu_f; \mu. + \beta_{at_f}, \Sigma_{at_f} + \Sigma_f) V(\mu_f).$$

For a given basis set of fixed states $\mu_f$, the majority of the right side can be computed once and used repeatedly during value iteration; essentially, the continuous-state MDP is converted to a new discrete-state MDP where the states are the fixed points.

At each timestep, the agent chooses the action that maximizes the estimate of its current value according to $Q_t$: $a = \operatorname{argmax}_a Q_t(s, a)$. The complete algorithm is shown in Algorithm 1.

## 3   LEARNING COMPLEXITY

In Section 4, we will analyze our algorithm in a family of MDPs with switching noisy offsets, and examine how many samples $N_{at}$ are necessary in order to produce a good policy. In particular, we prove it is probably approximately correct with a sample complexity ($N_{at}$) that scales polynomially with the number of dimensions in the state space.

When analyzing the performance of an RL algorithm $\mathcal{A}$, there are many potential criteria to use. In our work, we will focus predominantly on sample complexity with a brief mention of computational complexity. Computational complexity refers to the number of operations executed by the algorithm for each step taken by the agent in the environment. We will follow Kakade (2003) and use *sample complexity* as shorthand for the *sample complexity of learning*. It is the number of timesteps at which the algorithm, when viewed as a non-stationary policy $\pi$, is not $\epsilon$-optimal at the current state; that is, $Q^*(s, a) - Q^\pi(s, a) > \epsilon$ where $Q^*$ is the optimal state-action value function and $Q^\pi$ is the state–action value function of the non-stationary policy $\pi$. Following Strehl et al. (2006), we are interested in showing, for a given $\epsilon$ and $\delta$, that with probability at least $1 - \delta$ the sample complexity of the algorithm is less than or equal to a polynomial function of MDP parameters. Note that we only consider the number of samples to ensure the algorithm will learn and execute a near-optimal policy with high probability. As the agent acts in the world, it may be unlucky and experience a series of state transitions that poorly reflect the true dynamics, due to noise.

We will follow the lead of a recent and related continuous-state reinforcement-learning algorithm by Strehl and Littman (2008), and use the framework of Strehl et al. (2006). Strehl et al. (2006) defined an algorithm to be greedy if it chooses its action to be the one that maximizes the value of the current state $s$ ($a = \operatorname{argmax}_{a \in A} Q(s, a)$). Their paper's main result goes as follows: let $\mathcal{A}(\epsilon, \delta)$ denote a greedy learning algorithm. Maintain a list $K$ of "known" state–action pairs. At each new timestep, this list stays the same unless during that timestep a new state–action pair becomes known. MDP $M_K$ is the known state–action MDP (where the construction is essentially the same as described earlier, except that the reward and transition functions are the same as the original MDP for known state–action pairs) and $\pi$ is the greedy policy with respect to the current value function, $Q_t$. Assume that $\epsilon$ and $\delta$ are given and the following 3 conditions hold for all states, actions and timesteps:

1. $Q^*(s, a) - Q_t(s, a) \leq \epsilon$.

2. $V_t(s) - V_{M_K}^{\pi_t}(s) \leq \epsilon$.

3. The total number of times the agent visits a state–action tuple that is not in $K$ is bounded by $\zeta(\epsilon, \delta)$ (the *learning complexity*).

Then, Strehl et al. (2006) show on any MDP $M$, $\mathcal{A}(\epsilon, \delta)$ will follow a $4\epsilon$-optimal policy from its initial state on all but $N_{total}$ timesteps with probability at least $1 - 2\delta$, where $N_{total}$ is a polynomial in the problem's parameters $(\zeta(\epsilon, \delta), \frac{1}{\epsilon}, \frac{1}{\delta}, \frac{1}{1-\gamma})$.

The majority of our analysis will focus on showing that our algorithm fulfills these three criteria. In our approach,

we will define the known state–action pairs to be all those state–actions for which the type–action pair $\langle t(s), a \rangle$ is known.

Before we commence, we first briefly give some intuition for the above three criteria and describe how we will proceed in proving our algorithm satisfies them. Together, the first and second criteria can be interpreted as saying that the algorithm should produce accurate value estimates of the all state-action pairs in the known MDP, and that it should be optimistic about the values of all state–action pairs. The first criterion is more challenging to demonstrate. To show our estimates of known state–action pairs are close to their real values, we must consider two potential sources of error that could prevent it. The first is that the model dynamics are only estimated from the samples experienced, and so the dynamics model estimates may deviate from the true dynamics. In Proposition 4.1 and Lemmas 4.2, 4.3, and 4.4, we bound the number of samples necessary to ensure the dynamics model parameter estimates are close to the true dynamics. The second source of error comes from solving the MDP. We cannot currently perform exact optimal planning for these continuous-state noisy offset MDPs, and therefore we use approximate planning. In Section 4.2, we bound the error it introduces. We then combine these results in Lemma 4.5 to bound the error between our estimate of the value of the known-state MDP and the true optimal values. Theorem 4.6 uses this result to prove the algorithm is probably approximately correct with a sample complexity that scales polynomially in the problem parameters, including the state-space dimension.

Note that our use of an approximate planner is a departure from most related work on PAC RL. Existing work typically assumes the existence of a planning oracle for choosing actions given the estimated model.

To ensure fitted value iteration produces highly accurate results, our algorithm's worst-case computational complexity is exponential in the number of state dimensions. While this fact prevents it from being theoretically computationally efficient, our experimental results demonstrate our algorithm performs well compared to related approaches in a real-life robot problem.

## 4 ANALYSIS

This section provides a formal analysis of Algorithm 1. For simplicity, it assumes a diagonal covariance matrix for the noise model: $\Sigma = \text{diag}(\sigma_1^2, \sigma_2^2, \cdots, \sigma_{N_{dim}}^2)$. We believe it is possible to extend the analysis to the general covariance matrices, and leave it for future work. We also assume that the absolute values of the components in $\beta_{at}$ and $\Sigma_{at}$ are upper bounded by some known constants, $B_\beta$ and $B_\sigma$, respectively. This assumption is often true in practice. We denote by $|D|$ the determinant of matrix $D$. Due to space limitations some details will be omitted in our analysis: please see Brunskill et al. (2008) for full proofs.

### 4.1 MODEL ACCURACY

We first establish the distance between two dynamics models with different parameters. It will be important for analyzing the potential difference in expected received reward between a MDP with the true dynamics model and an estimated (from the data) dynamics model. Following Abbeel and Ng (2005), we use the variational distance

$$d_{var}(P(x), Q(x)) = \frac{1}{2} \int_{\mathcal{X}} |P(x) - Q(x)| dx. \quad (7)$$

**Proposition 4.1** *Assume that both $\Sigma_1$ and $\Sigma_2$ are diagonal matrices and let $\sigma_{\min}$ be the minimum standard deviation along any of the dimensions. Also, assume without loss of generality that $|\Sigma_1| \leq |\Sigma_2|$. Then,*

$$d_{var}(\mathcal{N}(s'|\beta_1 + s, \Sigma_1), \mathcal{N}(s'|\beta_2 + s, \Sigma_2))$$
$$\leq 1 - \left(\prod_{i=1}^{N_{dim}} \frac{\min[\sigma_{1i}^2, \sigma_{2i}^2]}{\sigma_{2i}^2}\right)^{0.5} + \frac{||\beta_2 - \beta_1||_2}{\sqrt{(2\pi)}\sigma_{\min}},$$

*where $\sigma_{ki}^2$ is the $i$-th diagonal component of $\Sigma_k$.*

**Proof**

$$d_{var}(\mathcal{N}(s'|\beta_1 + s, \Sigma_1), \mathcal{N}(s'|\beta_2 + s, \Sigma_2))$$
$$= \frac{1}{2} \int_{s'} |\mathcal{N}(s'|\beta_1 + s, \Sigma_1) - \mathcal{N}(s'|\beta_2 + s, \Sigma_2)| ds'$$
$$= \frac{1}{2} \int_{s'} |\mathcal{N}(s'|\beta_1 + s, \Sigma_1) - \mathcal{N}(s'|\beta_2 + s, \Sigma_1) +$$
$$\quad \mathcal{N}(s'|\beta_2 + s, \Sigma_1) - \mathcal{N}(s'|\beta_2 + s, \Sigma_2)| ds',$$

where we have simply added and subtracted the same term. Using the triangle inequality, we can split the expression into two terms:

$$d_{var}(\mathcal{N}(s'|\beta_1 + s, \Sigma_1), \mathcal{N}(s'|\beta_2 + s, \Sigma_2))$$
$$\leq \frac{1}{2} \int_{s'} |\mathcal{N}(s'|\beta_1 + s, \Sigma_1) - \mathcal{N}(s'|\beta_2 + s, \Sigma_1)| ds'$$
$$+ \frac{1}{2} \int_{s'} |\mathcal{N}(s'|\beta_2 + s, \Sigma_1) - \mathcal{N}(s'|\beta_2 + s, \Sigma_2)| ds',$$

one where the means are the same and the variances are different, and one where the variances are the same and the means are different. The second term is summing all the area between the lines defining the two Gaussians (which are centered at the same mean). An alternate way to think about computing this area is to take the sum of the area under the two Gaussians (which is simply 2) and subtract off two times the area of the intersection, $D$, between them:

$$\frac{1}{2}(2 - 2D) = 1 - D. \quad (8)$$

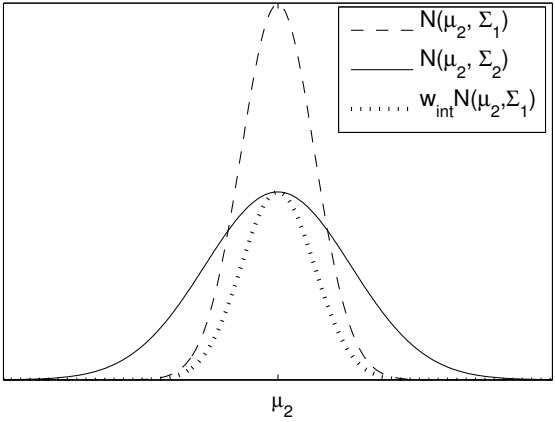

Figure 1: Two Gaussians with identical means and different variances, and a new weighted Gaussian that lies entirely inside their intersection.

To upperbound this term, we would like to find a lower bound on the area of intersection between these two Gaussians, $D$. We can construct a new weighted Gaussian that lies entirely within the intersection area and has the same mean as the two Gaussians ($\beta_2 + s$) (see Figure 1 for a one-dimensional example). We can set the covariance of this new Gaussian by setting its variance along each dimension $i$ to be the smaller of the two Gaussians' variances: $\sigma^2_{int,i} = \min[\sigma^2_{i1}, \sigma^2_{i2}]$. We then determine the weight on the Gaussian $w_{int}$ by requiring that its height at the mean be no more than the smaller of the two Gaussians. Since we have assumed that $|\Sigma_1| \leq |\Sigma_2|$, then the height at the mean of the smaller Gaussian is simply $1/((2\pi)^{N_{dim}/2}|\Sigma_2|^{0.5})$. Therefore, we can set $w_{int}$ as

$$w_{int}\mathcal{N}_{int}(0, \Sigma_{int}) = \frac{1}{(2\pi)^{N_{dim}/2}|\Sigma_2|^{0.5}}.$$

Solving for $w_{int}$, we get

$$w_{int} = \left( \prod_{i=1}^{N_{dim}} \frac{\min[\sigma^2_{1i}, \sigma^2_{2i}]}{\sigma^2_{2i}} \right)^{0.5}.$$

This weighted Gaussian always lies within the intersection region and therefore the $D$ is at least

$$\int w_{int}\mathcal{N}_{int}(s'|\beta_2 + s, \Sigma_{int})ds' = w_{int}.$$

Substituting this expression back into Equation 8,

$$\frac{1}{2}\int_{s'} |\mathcal{N}(s'|\beta_2 + s, \Sigma_1) - \mathcal{N}(s'|\beta_2 + s, \Sigma_2)|\,ds'$$
$$\leq\; 1 - w_{int} \;=\; 1 - \left( \prod_{i=1}^{N_{dim}} \frac{\min[\sigma^2_{1i}, \sigma^2_{2i}]}{\sigma^2_{2i}} \right)^{0.5}. \quad (9)$$

Next, consider the first term in Equation 8, which looks at the difference between two Gaussians with different means and identical variances. From Abbeel and Ng (2005) (Proposition 7), this expression is upperbounded by

$$\frac{||\beta_2 + s - (\beta_1 + s)||_2}{\sqrt{2\pi}\sigma_{\min}} = \frac{||\beta_2 + \beta_1||_2}{\sqrt{2\pi}\sigma_{\min}}. \quad (10)$$

Combining Equations 10 and 9 gives the desired result. □

Note this function is 0 when the means and the variances are the same, as one would hope.[3]

We next seek to determine the number of samples necessary to ensure that $d_{var}$ is tightly bounded when evaluated at the estimated model parameters and the true model parameters. Let us first define "good" samples as those for which $\|s' - s\|_\infty < B$ for some given $B > 0$. The value of $B$ will be specified later.

**Lemma 4.2** *Given any $\epsilon, \delta > 0$, define $T_\beta = \frac{2N_{dim}B^2}{\epsilon^2}\ln\frac{6N_{dim}}{\delta}$. If there are $T_\beta$ good transition samples $(s, a, s')$, then with probability at least $1 - \frac{\delta}{3}$, the estimated offset parameter $\tilde{\beta}$, computed by Equation 3, deviates from the true offset parameter $\beta^*$ by at most $\epsilon$; formally, $\Pr(\|\tilde{\beta} - \beta^*\|_2 \leq \epsilon) \geq 1 - \frac{\delta}{3}$.*

**Proof** : (Sketch) The proof rests on an application of Hoeffding's inequality (Hoeffding, 1963). □

We next analyze the number of samples needed to estimate the variance accurately.

**Lemma 4.3** *Assume $\|\tilde{\beta} - \beta\|_2 \leq \epsilon$. Given any $\epsilon, \delta > 0$, define $T_\sigma = \frac{8B^4}{\epsilon - \epsilon^2}\ln\frac{6N_{dim}}{\delta}$. If there are $T_\sigma$ good transition samples $(s, a, s')$, then with probability at least $1 - \frac{\delta}{3}$, the estimated variance parameter $\tilde{\sigma}_i^2$, computed by Equation 4, deviates from the true variance parameter $\sigma_i^2$ by at most $\epsilon$ for every dimension $i$; formally, $\Pr(\max_i |\tilde{\sigma}_i^2 - \sigma_i^2| \leq \epsilon) \geq 1 - \frac{\delta}{3}$.*

**Proof** : (Sketch) The proof first relates the estimate of the variance using the current estimate of the offset parameter $\beta$ to the variance around the true offset parameter, and then bounds this error using Hoeffding's inequality and a union bound. □

These two lemmas provide us with an estimate of how many good samples are necessary to achieve, with high probability, accurate estimates of the dynamics model parameters for every type–action pair. One additional lemma is needed to bound how many samples must be collected until enough such good samples are obtained.

---

[3]The true $d_{var}$ is upper bounded by 1, whereas this expression can go higher, so it is overly pessimistic when the difference between the two Gaussians' parameters is large, but increasingly accurate as their difference goes to 0. Since we need to estimate the parameters fairly precisely, we are more concerned with this second case.

**Lemma 4.4** *Let $T$ be the number of observed samples before $T_0 = \max\{T_\beta, T_\sigma\}$ good samples are collected. Then, $\Pr(T > \frac{\delta T_0}{\delta - 3N_{dim}p_0}) < \frac{\delta}{3}$, where $p_0 = \sqrt{\frac{8}{\pi} \frac{B_\sigma^3}{(B-B_\beta)^3}}$. Here, setting $B > B_\beta + \sqrt[6]{\frac{72N_{dim}^2}{\pi\delta^2}} B_\sigma$ ensures $\delta > 3N_{dim}p_0$.*

**Proof**: It follows from a union bound that $\Pr(\|s' - s\|_\infty > B) \leq N_{dim}\Pr(|s'_i - s_i| > B)$ for all $i$. We will show that $\Pr(|s'_i - s_i| > B)$ is small. Let $\varphi(x)$ and $\Phi(x)$ be the probability density function and cumulative distribution function of the standard Gaussian distribution, respectively. Then,

$$\Pr(s'_i - s_i > B) = 1 - \Phi\left(\frac{B - \beta_i^*}{\sigma_i}\right)$$
$$\leq \frac{1}{\sqrt{2\pi}} \exp\left(-\frac{(B - \beta_i^*)^2}{2\sigma_i^2}\right) \frac{1}{\frac{B - \beta_i^*}{\sigma_i}}$$
$$= \frac{\sigma_i}{\sqrt{2\pi}(B - \beta_i^*)} \exp\left(-\frac{(B - \beta_i^*)^2}{2\sigma_i^2}\right),$$

where the first equality follows from the definition, and the inequality follows from the fact that $1 - \Phi(y) < \frac{\varphi(y)}{y}$ when $y > 0$. Now, we can apply the inequality $e^{-x} < \frac{1}{1+x}$ to obtain

$$\Pr(s'_i - s_i > B) \leq \frac{\sigma_i}{\sqrt{2\pi}(B - \beta_i^*)} \cdot \frac{1}{1 + \frac{(B - \beta_i^*)^2}{2\sigma_i^2}}$$
$$< \sqrt{\frac{2}{\pi}} \frac{\sigma_i^3}{(B - \beta_i^*)^3} \leq \sqrt{\frac{2}{\pi}} \frac{B_\sigma^3}{(B - B_\beta)^3}.$$

Similarly, we may upperbound $\Pr(s'_i - s_i < -B)$ and thus, $\Pr(|s'_i - s_i| > B) < p_0$, where $p_0$ is given in the lemma statement.

Now, return to the full multivariate case:

$$\Pr(\|s' - s\|_\infty > B) \leq N_{dim}p_0 = \sqrt{\frac{8}{\pi} \frac{B_\sigma^3 N_{dim}}{(B - B_\beta)^3}}.$$

This inequality indicates that every sample is a "bad" sample with probability at most $N_{dim}p_0$. Given $T$ i.i.d. samples, let $N(T)$ be the number of bad samples. Our estimation algorithm fails to have $T_0$ good samples if and only if $N(T) > T - T_0$. By Markov's inequality,

$$\Pr(N(T) > T - T_0) \leq \frac{\mathbf{E}[N(T)]}{T - T_0} < \frac{TN_{dim}p_0}{T - T_0}.$$

Solving for $T$ by letting the last expression equal $\frac{\delta}{3}$ gives $T = \frac{\delta T_0}{\delta - 3N_{dim}p_0}$. We can obtain the minimum value of $B$ by solving $3N_{dim}p_0 = \delta$ for $B$. □

Combining these results with Lemmas 4.2 and 4.3 gives a condition on the minimum number of samples necessary to ensure, with high probability, the estimated parameters of a particular type–action dynamics model are close to the true parameters:

$$T = \max\{T_\beta, T_\sigma\} = O\left(\frac{N_{dim}B^4}{\epsilon^2} \ln \frac{N_{dim}}{\delta}\right).$$

## 4.2 PLANNING ERROR

We next bound the error between the value function found by solving our particular continuous-state Markov decision process using fitted value iteration compared to the optimal value function $V^*$. Recall that by performing FVI, we are essentially mapping the original MDP to a new finite-state MDP where the states are the chosen fixed points.

Under a set of four assumptions, Chow and Tsitsiklis (1991) proved that the optimal value function $V_\varepsilon$ of a discrete-state MDP formed by discretizing a continuous-state MDP into O($\varepsilon$)-length (per dimension)[4] grid cells is an $\varepsilon$-close approximation of the optimal continuous-state MDP value function $V^*$:

$$\|V_\varepsilon - V^*\| \leq \varepsilon.$$

The first two assumptions used to prove the above result are that the reward function and probability distribution are Lipschitz-continuous. In our work, the reward function is assumed to be given so this condition is a prior condition on the problem specification. Our probability distributions are Gaussian distributions that are Lipschitz-continuous so the second condition holds. The third key assumption is that the dynamics probabilities represent a true probablity measure that sums to 1 ($\int'_s p(s'|s,a) = 1$), though the authors show that this assumption can be relaxed to $\int'_s p(s'|s,a) \leq 1$ and the main results still hold. In our work, our dynamics models are defined to be true probability models. Chow and Tsitsiklis's final assumption is that there is a bounded difference between any two controls: in our case we consider only finite controls, so this property holds directly.

In summary, assuming the reward model fulfills the first assumption, our framework satisfies all four assumptions made by Chow and Tsitsiklis. Therefore, by selecting fixed grid points at a regular spacing of O($\epsilon_{FVI}$) in each dimension (letting $\varepsilon = \epsilon_{FVI}$), we can ensure that $\|\tilde{V}_{FVI} - V^*\|_\infty$ is at most $\epsilon_{FVI}$ where $\tilde{V}_{FVI}$ is the FVI optimal value function.

---
[4]More specifically, the grid spacing $h_g$ must satisfy $h_g \leq \frac{(1-\gamma)^2 \varepsilon}{K_1 + 2KK_2}$ and $h_g \leq \frac{1}{2K}$ where $K$ is the larger of the Lipschitz constants arising from the assumptions discussed in the text, and $K_1$ and $K_2$ are constants discussed in Chow and Tsitsiklis (1991). For small $\varepsilon$ any $h_g$ satisfying the first condition will automatically satisfy the second condition.

### 4.3 APPROXIMATE REINFORCEMENT LEARNING

The next lemma relates the accuracy in the dynamics model parameters, and the error induced by approximate planning, to the value function of two MDPs. The proof strongly parallels a similar Simulation Lemma in recent work by Strehl and Littman (2008).

**Lemma 4.5** *Let $M_1 = \langle S, A, p_1(s'|s,a), R, \gamma \rangle$ and $M_2 = \langle S, A, p_2(s'|s,a), R, \gamma \rangle$ be two MDPs[5] with dynamics as characterized in Equation 1 and non-negative rewards bounded above by 1. Assume $\frac{||\beta_1 - \beta_2||_2}{\sqrt{2\pi}\sigma_{\min}} \leq F_1$ and $|1 - \frac{|\Sigma_1|^{0.5}}{|\Sigma_2|^{0.5}}| \leq F_2$. Also assume that the difference between the value function $\tilde{V}$ obtained by fitted value iteration (FVI) compared to the optimal value function $V^*$, $||\tilde{V} - V^*||_\infty$ is at most $F_3$. Let $\pi$ be a policy that can be applied to both $M_1$ and $M_2$. Then, for any $0 < \epsilon \leq V_{\max}$ and stationary policy $\pi$, if $F_1 = O(\frac{(1-\gamma)^2 \epsilon}{\gamma})$ $F_2 = O(\frac{\epsilon(1-\gamma)^2}{\gamma})$, and $F_3 = O(\frac{\epsilon(1-\gamma)}{\gamma})$, then for all states $s$ and actions $a$, $|Q_1^\pi(s,a) - \tilde{Q}_2^\pi(s,a)| \leq \epsilon$, where $\tilde{Q}_2^\pi$ denotes the state-action value obtained by performing FVI on MDP $M_2$ and $Q_1^\pi$ denotes the true state-action value for MDP $M_1$ for policy $\pi$.*

**Proof**: (Sketch) We analyze the norm between $Q_1^\pi(s,a)$ and $\tilde{Q}_2^\pi(s,a)$ by re-expressing each in terms of its respective Bellman operator. The main idea is to break the norm up into a difference between the values due to the different dynamics ($p_1(s'|s,a)$ and $p_2(s'|s,a)$) and a difference due to using fitted value iteration to approximately solve for the values versus an exact solution. We use the triangle inequality to separate these terms and then bound each term separately, using the results from the prior sections. □

### 4.4 APPROXIMATELY OPTIMAL REINFORCEMENT LEARNING

**Theorem 4.6** *For any given $\delta$ and $\epsilon$ in a continuous-state noisy offset dynamics MDP with $N_T$ types where the variance along each dimension of all the dynamics models is bounded by $[\sigma_{\min}^2, B_\sigma^2]$ and the offset parameter is bounded by $|\beta_i| < B_\beta$ on all but $N_{total}$ timesteps, our algorithm will follow a $4\epsilon$-optimal policy from its current state with probability at least $1 - 2\delta$, where $N_{total}$ is polynomial in the problem parameters $(N_{dim}, |A|, N_T, \frac{1}{\epsilon}, \frac{1}{\delta}, \frac{1}{1-\gamma}, \frac{1}{\sigma_{\min}}, B_\beta, B_\sigma)$.*

**Proof**: We demonstrate that our algorithm fulfills the three criteria outlined earlier. We omit details due to space considerations, but it can be shown using the results of the prior sections that after $N_{at} = O\left(\frac{N_{dim}^3 B^4 \gamma^2}{\sigma_{\min}^4 (1-\gamma)^4 \epsilon^2}\right)$ samples, with probability $1 - \delta$, the errors $||\beta_1 - \beta_2||_2$, and for each state dimension $i$, $|\sigma_i^2 - \tilde{\sigma}_i^2|$ will be $O((1-\gamma)^2 \epsilon)$. We also chose the spacing of our fixed grid points such that $\frac{\epsilon_{FVI}\gamma}{1-\gamma} \leq \frac{\epsilon}{2}$. Then, the Simulation Lemma (4.5) guarantees that the approximate value of our known state MDP solved using FVI is $\epsilon$-close to the optimal value of the known state MDP with the true dynamics parameters $||\tilde{V}_{\tilde{K}}^\pi - V_K^\pi||_\infty \leq \epsilon$. All unknown type–action pairs that have not yet been experienced $N_M$ times are considered to be unknown and their value is set to $V_{\max}$. So, Conditions (1) and (2) (Strehl et al., 2006) hold. The third condition limits the number of times the algorithm may experience an unknown type–action tuple. Since there are a finite number of types and actions, this quantity is bounded above by $N_{at}N_T|A|$, which is a polynomial in the problem parameters $(N_{dim}, |A|, N_T, \frac{1}{\epsilon}, \frac{1}{\delta}, \frac{1}{1-\gamma}, \frac{1}{\sigma_{\min}}, B_\beta, B_\sigma)$. Therefore, our algorithm fulfills the three criteria laid out and the result follows. □

## 5 EXPERIMENT

To examine the performance of our algorithm, we performed experiments in a real-life robotic environment involving a navigation task where a robotic car must traverse multiple surface types to reach a goal location. Our experiments seek to demonstrate both that our dynamics models provide a sufficiently good representation of real-world dynamics to allow our algorithm to learn good policies, and that our algorithm is computationally tenable. We demonstrate the second quality by comparing to Leffler et al. (2007)'s RAM-Rmax algorithm, a provably efficient RL algorithm for learning in discrete-state worlds with types. The authors demonstrated that, by explicitly representing the types, they could get a significant learning speedup compared to Rmax, which learns a separate dynamics model for each state. The RAM-Rmax algorithm represents the dynamics model using a list of possible next outcomes for a given type. Our approach instead assumes a fixed parametric distribution that automatically constrains the size of the representation.

### 5.1 EXPERIMENTAL SETUP

For our experiment, we ran a LEGO® Mindstorms NXT robot on a multi-surface environment. A tracking pattern was placed on the top of the robot and an overhead camera was used to determine the robot's current position and orientation. The domain, shown in Figure 2, consisted of two types: rocks embedded in wax and a carpeted area. The goal was for the agent to begin in the start location (indicated in the figure by an arrow) and end in the goal without going outside the environmental boundaries. The rewards were $-1$ for going out of bounds, $+1$ for reaching the goal, and $-0.01$ for taking an action. Reaching the goal and go-ing out of bounds ended the episode and resulted in the agent getting moved back to the start location.

---
[5] For simplicity we present the results here without reference to types. In practice, each dynamics parameter would be subscripted by its associated MDP, type, and action.

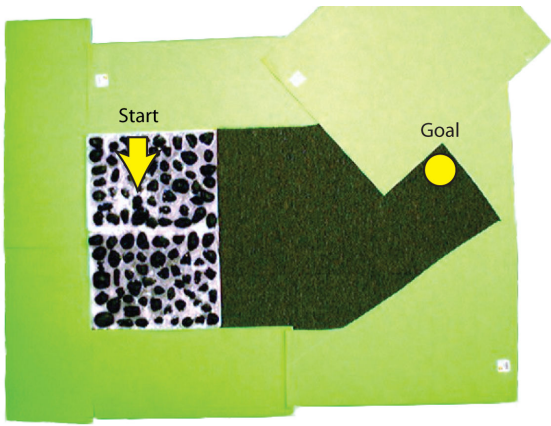

Figure 2: Image of the environment. The start location and orientation is marked with an arrow. The goal location is indicated by the circle.

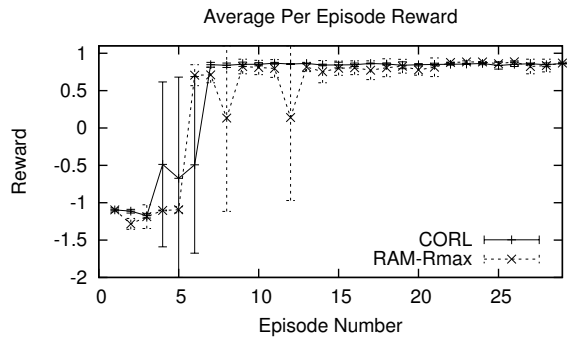

Figure 3: Reward received by algorithms averaged over three runs. Error bars show one standard deviation.

One difficulty of this environment is the difference in dynamics models. Due to the close proximity of the goal to the boundary, the agent needs an accurate dynamics model to reliably reach the goal. To make this task even more difficult, the actions were limited to going forward, turning left, and turning right. Without the ability to move backwards, the robot needed to approach the goal accurately to avoid falling out of bounds. A robot with an inaccurate transition model would be likely to judge this task as impossible.

For the experiments, we compared our algorithm ("CORL") and the RAM-Rmax algorithm ("RAM"). The fixed points for the fitted value iteration portion of our algorithm were set to the discretized points of the RAM-Rmax algorithm. Both algorithms used an EDISON image segmentation system to uniquely identify the current surface type. The reward function was provided to both algorithms.

The state space is three dimensional: $x$, $y$, and orientation. Our algorithm implementation for this domain used a full covariance matrix to model the dynamic's variance model. For the RAM-Rmax agent, the world was discretized to a forty-by-thirty-by-ten state space. In our algorithm, we used a function approximator of a weighted sum of Gaussians, as described in Section 2.2. We used the same number of Gaussians to represent the value function as the size of the state space used in the discretized algorithm, and placed these fixed Gaussians at the same locations. The variance over the $x$ and $y$ variables was independent of each other and of orientation, and was set to be 16. To average orientation vectors correctly (so that $-180°$ degrees and $180°$ do not average to 0) we converted orientations $\theta$ to a Cartesian coordinate representation $x_\theta = \cos(\theta), y_\theta = \sin(\theta)$. The variance over these two was set to be 9 for each variable (with zero covariance). For our algorithm and the RAM-Rmax algorithm, the value of $N_{at}$ was set to four and five, respectively, which was determined after informal experimentation. The discount factor was set to 1.

### 5.2 RESULTS

Figure 3 shows the average reward with standard deviation for each of the algorithms over three runs. Both algorithms are able to receive near-optimal reward on a consistent basis choosing similar paths to the goal. Our dynamics representation is sufficient to allow our algorithm to learn well in this real-life environment.

Examining the learned dynamics model parameters revealed that the dynamics model variances learned for the rocks were larger than the learned variance for carpet for certain actions. Naturally, an important question is whether modeling the differences in the dynamics models is necessary in order to achieve good performance: in other words, could the robot perform as well by modeling the terrain as a single type? Prior work by RAM-Rmax on a similar task compared using two types to one, and found that two types did result in a better learned policy(Leffler et al., 2008). This finding suggests that using multiple types to represent this environment provides measurable benefits.

In addition, by using a fixed parametric representation, the computational time per episode of our algorithm is roughly constant. In the implementation of RAM-Rmax, the computational time grew with the number of episodes due to its dynamics model representation: however, this difficulty could be ameliorated by maintaining a finite list of potential dynamics transitions. Nonetheless, these results suggest that our algorithm is computationally competitive with existing approaches to handle domains with typed dynamics.

In summary, the results on this task are encouraging since they indicate our algorithm can quickly and efficiently learn a good policy in a real-world environment with switching noisy offset dynamics.

# 6 CONCLUSION

We have presented a new reinforcement-learning algorithm for handling continuous-state typed worlds where the dynamics can be modeled as a noisy offset. In this work, we have assumed that the state types are fully observable. This assumption is likely to be realistic for certain domains, such as when types correspond to the slope of an outdoor environment in which contour maps are available. In other cases, it might be useful to model the type as a hidden variable, and receive estimates of it through the agent's sensors. Such a scenario is beyond the scope of this paper but would be interesting future work.

In conclusion, we proved that when the noise covariance matrix is diagonal, the algorithm is probably approximately correct with a sample complexity that scales polynomially with the MDP parameters, including the state-space dimension. We also demonstrated that in some scenarios these dynamics representations can provide a sufficiently good approximation of real-world dynamics to enable a good policy to be learned by demonstrating the success of our algorithm in a small robotic experiment.

**Acknowledgements**

B. Leffler, L. Li and M. Littman were partially supported by NSF DGE 0549115, a DARPA transfer learning grant and a National Science Foundation (NSF) Division of Information and Intelligent Systems (IIS) grant. E. Brunskill and N. Roy were supported by NSF IIS under Grant #0546467.